\documentclass[english]{llncs}
\usepackage[T1]{fontenc}
\usepackage[latin9]{inputenc}
\usepackage{multirow}
\usepackage{amsbsy}
\usepackage{amstext}
\usepackage{graphicx}
\usepackage[misc]{ifsym}

\makeatletter

\providecommand{\tabularnewline}{\\}

\usepackage{helvet}
\usepackage{courier}
\usepackage{scrextend}
\usepackage{hyperref}
\usepackage{babel}
\usepackage{multibib}
\newcites{append}{Appendix References}
\usepackage{caption}

\begin{document}

\author{Shaohua Li \inst{1,2}$^\text{(\Letter)}$ \and Yong Liu \inst{1}  \and Xiuchao Sui \inst{2} \and Cheng Chen \inst{1} \and \\
Gabriel Tjio \inst{1} \and Daniel Shu Wei Ting \inst{3} \and Rick Siow Mong Goh \inst{1}}

% index{Li, Shaohua}
% index{Liu, Yong}
% index{Sui, Xiuchao}
% index{Chen, Cheng}
% index{Tjio, Gabriel}
% index{Ting, Daniel Shu Wei}
% index{Goh, Rick Siow Mong}

\institute{Institute of High Performance Computing, A*STAR, Singapore \\ \email{shaohua@gmail.com} \and Artificial Intelligence Initiative, A*STAR, Singapore \and Singapore Eye Research Institute}
% \author{Anonymous \\ Anonymous institute}

\makeatother

\title{Multi-Instance Multi-Scale CNN for Medical Image Classification}
\maketitle
\begin{abstract}
Deep learning for medical image classification faces three major challenges:
1) the number of annotated medical images for training are usually
small; 2) regions of interest (ROIs) are relatively small with unclear boundaries in the whole medical images, and may appear in arbitrary
positions across the $x,y$ (and also $z$ in 3D images) dimensions.
However often only labels of the whole images are annotated, and
localized ROIs are unavailable; and 3) ROIs in
medical images often appear in varying sizes (scales). We approach
these three challenges with a Multi-Instance Multi-Scale (MIMS) CNN:
1) We propose a multi-scale convolutional layer, which extracts patterns
of different receptive fields with a shared set of convolutional kernels, so that scale-invariant patterns are captured by this compact set of kernels.
As this layer contains only a small number of parameters, training
on small datasets becomes feasible; 2) We propose a ``top-$k$ pooling''
to aggregate the feature maps in varying scales from multiple spatial
dimensions, allowing the model to be trained using weak annotations
within the multiple instance learning (MIL) framework. Our method is shown to perform well on three classification tasks involving two 3D and two 2D medical image datasets.
\end{abstract}

\section{Introduction}

Training a convolutional neural network (CNN) from scratch demands
a massive amount of training images. Limited medical images encourage
people to do transfer learning, i.e., fine-tune 2D CNN models pretrained
on natural images \cite{finetune}. A key difference between medical images
and natural images is that, regions of interest (ROIs) are relatively small
with unclear boundaries in the whole medical images, and ROIs may appear multiple times in arbitrary positions across the $x,y$ (and also
$z$ in 3D images) dimensions. On the other hand, annotations for medical images are often ``weak'', in that only image-level annotations are available, and there are no localized ROIs. In this setting, we can view each ROI as an instance in a bag of all image patches, and the image-level classification falls within the Multiple-Instance Learning (MIL) framework \cite{mil-mammo,attnMIL,thoracic}.

Another challenge with medical images is that ROIs are
often \emph{scale-invariant}, i.e., visually similar patterns often
appear in varying sizes (scales). If approached with vanilla CNNs,
an excess number of convolutional kernels with varying receptive fields would be required for full coverage of these patterns, which have more parameters and demand more training data. Some previous works have attempted to learn scale-invariant patterns, for example \cite{oct-pyramid} adopted image pyramids, i.e. resizing input images into different scales, processing them with the same CNN and aggregating the outputs. However, our experiments show that image pyramids perform unstably across different datasets and consume much more computational resources than vanilla CNNs.

This paper aims to address all the challenges above in a holistic
framework. We propose two novel components: 1) a \emph{multi-scale
convolutional layer} (MSConv) that further processes feature maps
extracted from a pretrained CNN, aiming to capture scale-invariant
patterns with a shared set of kernels; 2) a \emph{top-$k$ pooling} scheme
that extracts and aggregates the highest activations from feature
maps in each convolutional channel (across multiple spatial dimensions
in varying scales), so that the model is able to be trained with image-level labels only.

The MSConv layer consists of a few resizing operators (with different output resolutions), and a shared set of convolutional kernels. First a pretrained CNN extracts feature maps from input images. Then the MSConv layer resizes them to different scales, and processes each scale with the same set of convolutional kernels. Given the varying scales of the feature maps, the convolutional kernels effectively have varying receptive fields, and therefore are able to detect scale-invariant patterns. As feature maps are much smaller than input images, the computation and memory overhead of the MSConv layer is insignificant.

The MSConv layer is inspired by ROI-pooling \cite{fast-rcnn}, and is closely related to Trident Network \cite{trident}. Trident Network uses shared convolutional kernels of different dilation rates to capture scale-invariant patterns. Its limitations include: 1) the receptive fields of dilated convolutions can only be integer multiples of the original receptive fields; 2) dilated convolutions may overlook prominent activations within a dilation interval. In contrast, the MSConv interpolates input feature maps to any desired sizes before convolution, so that the scales are more refined, and prominent activations are always retained for further convolution. \cite{siconv} proposed a similar idea of resizing the input multiple times before convolution and aggregating the resulting feature maps by max-pooling. However we observed that empirically, activations in larger scales tend to dominate smaller scales and effectively mask smaller scales. MSConv incorporates a batchnorm layer and a learnable weight for each scale to eliminate such biases. In addition, MSConv adopts multiple kernel sizes to capture patterns in more varying scales.

A core operation in an MIL framework is to aggregate features or predictions
from different instances (pattern occurrences). 
% As each output channel encodes a particular pattern, feature maps in the same output channel (across different dimensions in varying scales) have the same semantics and are grouped for aggregation. 
Intuitively, the most prominent patterns are usually also the most discriminative, and thus the highest activations could summarize a set of feature maps with the same semantics (i.e., in the same channel). In this regard, we propose a top-$k$ pooling scheme that selects the highest activations of a group of feature maps, and takes their weighted average as the aggregate feature for downstream processing. The top-$k$ pooling extends \cite{rankpool} with learnable pooling weights (instead of being specified by a hyperparameter as in \cite{rankpool}) and a learnable magnitude-normalization operator.

The MSConv layer and the top-$k$ pooling comprise our Multi-Instance
Multi-Scale (MIMS) CNN. To assess its performance, we
evaluated 12 methods on three classification tasks: 1) classifying Diabetic Macular Edema (DME) on three Retinal Optical Coherence Tomography (OCT) datasets (two sets of 3D images); 2) classifying Myopic Macular Degeneration (MMD) on a  2D fundus image dataset; and 3) classifying Microsatellite Instable (MSI) against microsatellite stable (MSS) tumors of colorectal cancer (CRC) patients on histology images. In most cases, MIMS-CNN achieved better accuracy than five baselines 
and six ablated models. Our experiments also verified that both the MSConv layer and top-$k$ pooling make important contributions. 
% These experiments demonstrate that our model works well on small weakly annotated datasets with low computational resources.

\setlength{\abovedisplayskip}{3pt} \setlength{\belowdisplayskip}{3pt}\setlength{\textfloatsep}{15pt}

\begin{figure}[!h]
\centering{}\includegraphics[trim=0bp 20bp 0bp 0bp,scale=0.33]{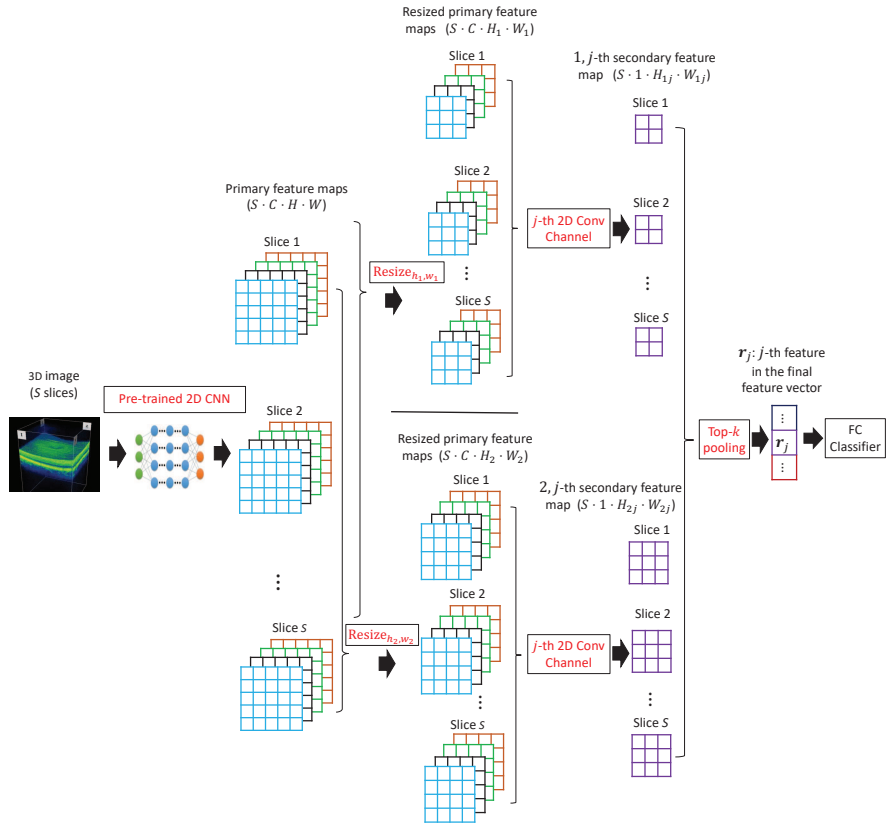}\caption{\label{fig:msconv}The Multi-Instance Multi-Scale CNN on a 3D input image. For clarity, only the $j$-th convolutional channel of the MSConv layer is shown.
%The pretrained CNN, MSConv layer and top-$k$ pooling 
%are highlighted in red. 
}
\end{figure}

\section{Multi-Instance Multi-Scale CNN}

The architecture of our Multi-Instance Multi-Scale CNN is illustrated
in Fig.~\ref{fig:msconv}. It consists of: 1) a pretrained 2D CNN to
extract primary feature maps, 2) a multi-scale convolutional (MSConv)
layer to extract scale-invariant secondary feature maps, 3) a top-$k$
pooling operator to aggregate secondary feature maps, and 4) a classifier.

\subsection{Multi-Scale Convolutional Layer\label{subsec:msconv}}

Due to limited training images, a common practice in medical image analysis is to extract image features using 2D CNNs pretrained on natural images. These features are referred as the \emph{primary feature maps}. Due to the domain gap between natural images and medical images, feeding primary feature maps directly into a classifier does not always yield good results. To bridge this domain gap, we propose to use an extra convolutional layer to extract more relevant features from primary feature maps. This layer produces the \emph{secondary feature maps.}

In order to capture scale-invariant ROIs, we resize the primary feature maps into different scales before convolution. Each scale corresponds to a separate pathway, and weights of the convolutional kernels in all pathways are tied. In effect, this convolutional layer has multiple receptive fields on the primary feature maps. We name this layer as a \emph{multi-scale convolutional (MSConv) layer}. 

More formally, let $x$ denote the primary feature maps, $\{F_{1},\cdots,F_{N}\}$
denote all the output channels of the MSConv layer\footnote{Each convolutional kernel yields multiple channels with different
semantics, so output channels are indexed separately, regardless of
whether they are from the same kernel.}, and $\{(h_{1},w_{1}),\cdots,(h_{m},w_{m})\}$ denote the scale factors of the heights and widths (typically $\frac{1}{4} <=h_{i}=w_{i} <= 2$) adopted by the $m$ resizing operators. The combination of the $i$-th scale and the $j$-th channel yields the $ij$-th secondary feature maps:
\begin{equation}
\boldsymbol{y}_{ij}=F_{j}\left(\textrm{Resize}_{h_{i},w_{i}}(\boldsymbol{x})\right),
\end{equation}
where in theory $\textrm{Resize}_{h_{i},w_{i}}(\cdot)$ could adopt any type
of interpolation, and our choice is bilinear interpolation.

For more flexibility, the convolutional kernels in MSConv could also have different kernel sizes. In a setting of $m$ resizing operators and $n$ different sizes of kernels, effectively the kernels have at most $m\times n$ different receptive fields. The multiple resizing operators and varying sizes of kernels complement each other and equip the CNN with scale-invariance.

Among $\{\boldsymbol{y}_{1j},\boldsymbol{y}_{2j},\cdots,\boldsymbol{y}_{mj}\}$, feature maps in larger scales contain more elements and tend to have more top $k$ activations, hence dominate the aggregate feature and effectively mask out the feature maps in smaller scales. In order to remove such biases, the feature maps in different
scales are passed through respective magnitude normalization operators. The magnitude normalization operator consists of a batchnorm operator $\textrm{BN}_{ij}$ and a learnable scalar multiplier $sw_{ij}$. The scalar multiplier $sw_{ij}$ adjusts the
importance of the $j$-th channel in the $i$-th scale, and is optimized with back-propagation. 

\begin{figure}
\begin{centering}
\includegraphics[trim=0bp 350bp 0bp 0bp,clip,scale=0.26]{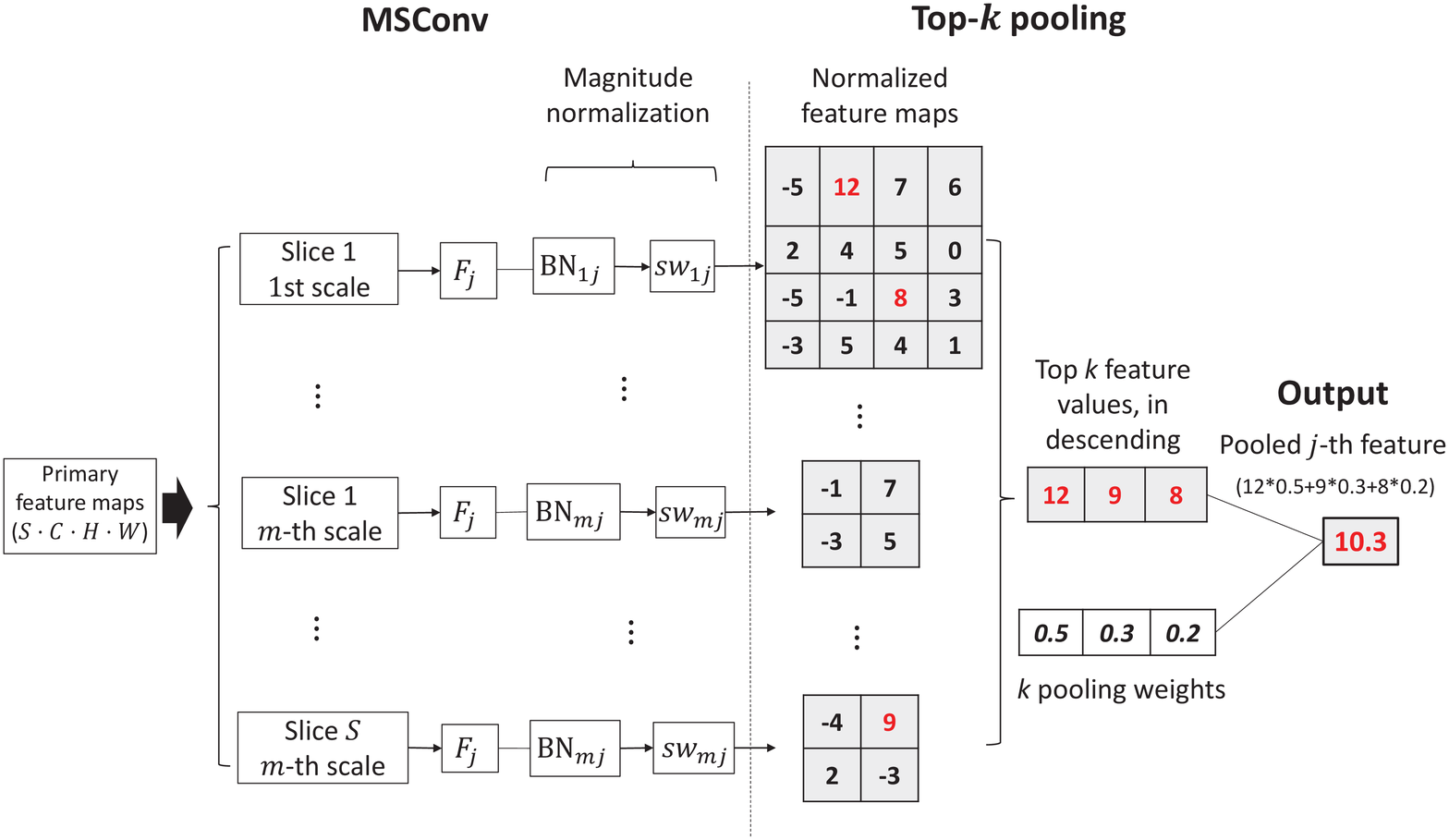}
\par\end{centering}
\caption{\label{fig:top-k-pooling} The MSConv and Top-$k$ pooling (on the $j$-th channel $F_j$ only) in $m$ scales.}
\end{figure}

The MSConv layer is illustrated in Fig.~\ref{fig:msconv} and the left side of Fig.~\ref{fig:top-k-pooling}.

\subsection{Top-$k$ Pooling\label{subsec:Top-k-Pooling}}

Multiple Instance Learning (MIL) views the whole image as a bag, and
each ROI as an instance in the bag. Most existing MIL works \cite{mil-mammo,thoracic} were \emph{instance-based
MIL}, i.e., they aggregate label predictions on instances to yield
a bag prediction. In contrast, \cite{attnMIL} adopted \emph{embedding-based MIL}, which aggregates features (embeddings) of instances to yield bag features, and then do classification on bag features. \cite{attnMIL} showed that embedding-based MIL methods outperformed instance-based MIL baselines. Here we propose a simple but effective \emph{top-$k$ pooling} scheme to aggregate the most prominent features across a few spatial dimensions, as a new embedding-based MIL aggregation scheme.

Top-$k$ pooling works as follows: given a set of feature maps with
the same semantics, we find the top $k$ highest activation values,
and take a weighted average of them as the aggregate feature value.
Intuitively, higher activation values are more important than lower
ones, and thus the pooling weight should decrease as the ranking goes lower. However it may be sub-optimal to specify the weights manually as did in \cite{rankpool}. Hence we adopt a data-driven approach to learn these weights automatically. More formally, given a set of feature maps $\{\boldsymbol{x}_{i}\}$, top-$k$ pooling aggregates them into a single value:
\begin{equation}
\textrm{Pool}_{k}(\{\boldsymbol{x}_{i}\})=\sum_{r=1}^{k}w_{r}a_{r},
\end{equation}
where $a_{1},\cdots,a_{k}$ are the highest $k$ activations within
$\{\boldsymbol{x}_{i}\}$, and $w_{1},\cdots,w_{k}$ are nonnegative
pooling weights to be learned, subject to
a normalization constraint $\sum_{r}w_{r}=1$. In practice, $w_{1},\cdots,w_{k}$
is initialized with exponentially decayed values, and then optimized
with back-propagation.

An important design choice in MIL is to choose the spatial dimensions to be pooled. Similar patterns, regardless of where they
appear, contain similar information for classification. Correspondingly,  features in the same channel could be pooled together. On 2D images, we choose to pool activations across the $x,y$-axes of the secondary feature maps, and on 3D images we choose to pool across the $x,y$ and $z$ (slices) axes. In addition, feature maps in the same channel but different scales (i.e., through different $\textrm{Resize}_{h_{i},w_{i}}(\cdot)$
and the same $F_{j}$) encode the same semantics and should be pooled
together. Eventually, all feature maps in the $j$-th channel,
$\{\boldsymbol{y}_{\cdot j}\}=\boldsymbol{y}_{1j},\boldsymbol{y}_{2j},\cdots,\boldsymbol{y}_{mj}$
are pooled into a single value $\textrm{Pool}_{k}(\{\boldsymbol{y}_{\cdot j}\})$.
Then following an $N$-channel MSConv layer, all feature maps will
be pooled into an $N$-dimensional feature vector to represent the
whole image. As typically $N<100$, the downstream FC layer doing
classification over this feature vector has only a small number of
parameters and less prone to overfitting.

Fig.\ref{fig:top-k-pooling} illustrates the
top-$k$ pooling being applied to the $j$-th channel feature maps
in $m$ scales.

\section{Experiments}

\subsection{Datasets}

Three classification tasks involving four datasets were used
for evaluation.

\emph{DME classification on OCT images}. The following two 3D datasets acquired by Singapore Eye Research Institute (SERI) were used:

1) \textbf{Cirrus} dataset: 339 3D OCT images (239 normal, 100 DME).
Each image has 128 slices in 512{*}1024. A 67-33\% training/test split
was used;

2) \textbf{Spectralis} dataset: 197 3D OCT images (60 normal, 137
DME). Each image has $25\sim31$ slices in 497{*}768. A 50-50\% training/test
split was used;

\emph{MMD classification on fundus images}:

3) \textbf{MMD} dataset (acquired by SERI): 19,272 2D images (11,924 healthy, 631 MMD) in 900{*}600. A 70-30\% training/test split was used.

\emph{MSI/MSS classification on CRC histology images}:

4) \textbf{CRC-MSI} dataset \cite{msi}: 93,408 2D training images (46,704 MSS, 46,704 MSI) in 224{*}224. 98,904 test images (70,569 MSS, 28,335 MSI) also in 224{*}224.

\subsection{Compared Methods}

MIMS-CNN, 5 baselines and 6 ablated models were compared. Unless specified, all methods used the ResNet-101 model (without FC) pretrained on ImageNet for feature extraction, and top-$k$ pooling ($k=5$) for feature aggregation.

\textbf{MI}-\textbf{Pre}. The ResNet feature maps are pooled by top-$k$ pooling and classified.

\textbf{Pyramid MI}-\textbf{Pre}. Input images are scaled to $\{\frac{i}{4}|i=2,3,4\}$
of original sizes, before being fed into the MI-Pre model.

\textbf{MI}-\textbf{Pre}-\textbf{Conv}. The ResNet feature maps are processed by an extra convolutional layer, and aggregated
by top-$k$ pooling before classification. It is almost
the same as the model in \cite{thoracic}, except that \cite{thoracic}
does patch-level classification and aggregates patch predictions to obtain image-level classification.

\textbf{MIMS}. The MSConv layer has 3 resizing operators that resize the primary feature maps to the following scales: $\{\frac{i}{4}|i=2,3,4\}$. Two groups of kernels of different sizes were used.

\textbf{MIMS-NoResizing}. It is an ablated MIMS-CNN with all resizing
operators removed. This is to evaluate the contribution of the resizing
operators.

\textbf{Pyramid MIMS}. It is an ablated MIMS-CNN with all resizing operators removed, and the multi-scaledness is pursued with input image pyramids of scales $\{\frac{i}{4}|i=2,3,4\}$. The MSConv kernels
is configured identically as above.

\textbf{MI-Pre-Trident} \cite{trident}. It extends MI-Pre-Conv with dilation factors ${1,2,3}$. 

\textbf{SI-CNN} \cite{siconv}. It is an ablated MIMS-CNN with the batchnorms and scalar multipliers $sw_{ij}$ removed from the MSConv layer.

\textbf{FeatPyra-4,5}. It is a feature pyramid network \cite{fpn} that extracts features from conv4\_x and conv5\_x in ResNet-101, processes each set of features with a respective convolutional layer, and classifies the aggregate features. 

\textbf{ResNet34-scratch}. It is a ResNet-34 model trained from scratch.

\textbf{MIMS-patchcls} and \textbf{MI-Pre-Conv-patchcls}. They are ablated MIMS and MI-Pre-Conv, respectively, evaluated on 3D OCT datasets. They classify each slice, and average slice predictions to obtain image-level classification.

\subsection{Results}

Table \ref{tab:scores} lists the AUROC scores (averaged over three independent runs) of the 12 methods on the four datasets. All methods with an extra convolutional layer on top of a pretrained model performed well. The benefits of using pretrained models are confirmed by the performance gap between ResNet34-scratch and others. The two image pyramid methods performed significantly worse on some datasets, although they consumed twice as much computational time and GPU memory as other methods. MIMS-CNN almost always outperformed other methods.
%The performance of the ablated MIMS-NoResizing revealed that the resizing operators contributed to the overall performance.
%\vspace{-1em}

\begin{table}
\centering{}%
\setlength{\tabcolsep}{10pt}
\begin{tabular}{c|c|c|c|c|c}
\hline 
Methods & Cirrus & \multirow{1}{*}{Spectralis} & MMD & CRC-MSI & Avg.\tabularnewline
\hline 
\hline 
MI-Pre          & 0.574 & 0.906 & 0.956 & 0.880 & 0.829\tabularnewline
\hline 
Pyramid MI-Pre  & 0.638 & 0.371 & 0.965 & 0.855 & 0.707\tabularnewline
\hline 
MI-Pre-Conv     & 0.972 & 0.990 & 0.961 & 0.870 & 0.948\tabularnewline
\hline 
MIMS-NoResizing & 0.956 & 0.975 & 0.961 & 0.879 & 0.942\tabularnewline
\hline 
Pyramid MIMS    & 0.848 & 0.881 & 0.966 & 0.673 & 0.842\tabularnewline
\hline 
MI-Pre-Trident         & 0.930 & \textbf{1.000} & 0.966 & 0.897 & 0.948\tabularnewline
\hline 
SI-CNN         & 0.983 & \textbf{1.000} & \textbf{0.972} & 0.880 & 0.959\tabularnewline
\hline 
FeatPyra-4,5    & 0.959 & 0.991 & 0.970 & 0.888 & 0.952\tabularnewline
\hline
ResNet34-scratch& 0.699 & 0.734 & 0.824 & 0.667 & 0.731\tabularnewline
\hline
MIMS            & \textbf{0.986} & \textbf{1.000} & \textbf{0.972} & \textbf{0.901} & \textbf{0.965}\tabularnewline
\hline 
MIMS-patchcls   & 0.874 & 0.722 & / & / & / \tabularnewline
\hline
MI-Pre-Conv-patchcls & 0.764 & 0.227 & / & / & / \tabularnewline
\hline
\end{tabular}\medskip{}
\caption{\label{tab:scores}Performance (in AUROC) of 12 methods on four image datasets.}
\end{table}

%\vspace{-1em}
The inferior performance of the two $*$-patchcls models demonstrated the advantages of top-$k$ pooling for MIL. To further investigate its effectiveness, we trained MIMS-CNN on Cirrus with six MIL aggregation schemes: average-pooling
(\textbf{mean}), max-pooling (\textbf{max}), top-$k$ pooling with
$k=2,3,4,5$, and an instance-based MIL scheme: max-pooling
over slice predictions (\textbf{max-inst}).

As can be seen in Table \ref{tab:pool-results}, the other three aggregation
schemes fell behind all top-$k$ schemes, and when $k$ increases,
the model tends to perform slightly better. It confirms that embedding-based
MIL outperforms instance-based MIL.

\begin{table}[t]
\centering{}%
\setlength{\tabcolsep}{5pt}
\begin{tabular}{c|c|c|c|c|c|c|c}
\hline 
Methods & mean & \multirow{1}{*}{max} & max-inst & $k=2$ & $k=3$ & $k=4$ & $k=5$\tabularnewline
\hline 
AUROC on Cirrus & 0.829 & 0.960 & 0.975 & 0.980 & 0.980 & 0.986 & 0.986 \tabularnewline
\hline 
\end{tabular}\medskip{}
\caption{\label{tab:pool-results}Performance of seven MIL aggregation schemes on the Cirrus dataset.}
\end{table}

\section{Conclusions}

Applying CNNs on medical images faces three challenges: datasets are of small sizes, annotations are often weak and ROIs are in varying scales.
We proposed a framework to address all these challenges. This framework
consists of two novel components: 1) a multi-scale convolutional layer on top of a pretrained
CNN to capture scale-invariant patterns, which contains only a small
number of parameters, 2) a top-$k$ pooling operator to aggregate
feature maps in varying scales across multiple spatial dimensions
to facilitate training with weak annotations within the Multiple Instance Learning framework. Our method has been validated on three classification tasks involving four image datasets.

\bibliographystyle{splncs04}
\bibliography{references}

\clearpage
\appendix

\begin{center}
\huge \textbf{Supplementary Material}
\end{center}
\section{Model Configurations and Training Settings}

For MIMS-CNN, on the two OCT datasets Cirrus and Spectralis, the convolutional
kernels were specified as $\{2(8),3(10)\}$ in ``kernel size (number
of output channels)'' pairs. On the MMD dataset, the convolutional
kernels were specified as $\{2(10),3(10)\}$. On the CRC-MSI dataset,
as the images are of smaller resolution, smaller convolutional kernels
$\{1(20),2(40)\}$ were adopted.

In all models, the underlying ResNet layers were fine-tuned to reduce
domain gaps between ImageNet and the training data. The learning rate
of the underlying layers was set as half of the top layers to reduce
overfitting.

To evaluate on OCT datasets, all models were first trained on Cirrus
for 4500 iterations. Then on Spectralis, the trained models were first
fine-tuned on the training set for 200 iterations, then evaluated
on the test sets. When training on the Cirrus and Spectralis datasets,
to increase data diversity, in each iteration $12\sim18$ slices were
randomly chosen to form a batch from the 30 central slices of the
input image.

On the CRC-MSI dataset, there is significant domain gap between the
training and test images. Hence 2\% of the original test images were
moved to the training set (these 2% are denoted as the \emph{tuning set}), to reduce this 
domain gap. In particular, all models were trained on the training set for one epoch (LR=0.01), 
and then fine-tuned on the tuning set for two epochs (LR=0.01, 0.004). 

When working on 3D images such as Cirrus and Spectralis, as the MSConv
layer only involves 2D convolutions, all slices in a 3D image can
be conveniently arranged as a 2D batch for faster processing.

\section{A Possible Explanation of Why Image Pyramid Failed}

We randomly selected 100 images from the CRC-MSI dataset \citeappend{msi},
scaled them with three scale factors $\{2,0.75,0.5\}$, and compared
the respective ResNet-101 features with those of the original images.
Their average Pearson correlations are listed in Table 1.

\begin{table}
\begin{centering}
\setlength{\tabcolsep}{0.5em}\begingroup\renewcommand*{\arraystretch}{1.2}%
\begin{tabular}{c|c}
\hline 
Scale & Pearson correlation\tabularnewline
\hline 
\hline 
2 & 0.261\tabularnewline
\hline 
0.75 & 0.451\tabularnewline
\hline 
0.5 & 0.257\tabularnewline
\hline 
\end{tabular}\bigskip{}
\par\end{centering}
\endgroup\caption{Pearson correlations of ResNet-101 features of the scaled images with
the original image features.}
\end{table}

\vspace{-2em}
One can see that the Pearson correlation dropped to around 0.26 when
the images were resized to half or double of their original sizes.
It indicates that ResNet captures very different features of them,
although they are the same images in different scales. This feature
decorrelation may prevent the downstream convolutional layer from
learning scale-invariant patterns with the same kernels. Instead,
the downstream convolutional layer needs to utilize more kernels to
memorize the different features at each scale, and thus more training
data is demanded. This observation may explain why image pyramid performed
inferiorly in our experiments.

\section{Gradient-based Localization and Visualization}

In practice, medical doctors are keen to understand how a Computer
Aided Diagnosis system reaches a certain diagnosis, so that they can make
reliable decisions. To this end, we adopt a variant of a gradient-based
visualization method \citeappend{grad-input,deepexplain,taylor}, making it reliably
locate and visualize suspicious slices and regions for doctors to
examine.

Suppose we have trained a classifier that classifies medical images.
To perform the visualization, we manually assign an interested class
to have loss $-1$, and other classes have loss 0, then do a backward
pass from the classifier to get the gradient at a pre-specified feature
map layer. When the input image is 3D, the classifier will first determine
which slices belong to the target class. For each of these slices,
the back-propagation based visualization method is applied to generate
a heatmap that quantifies the contributions of each pixel in the input
2D slice, and overlay the heatmap on the 2D slice as the output visualization
image.

\begin{figure}[th]
\centering{}\includegraphics[scale=0.25]{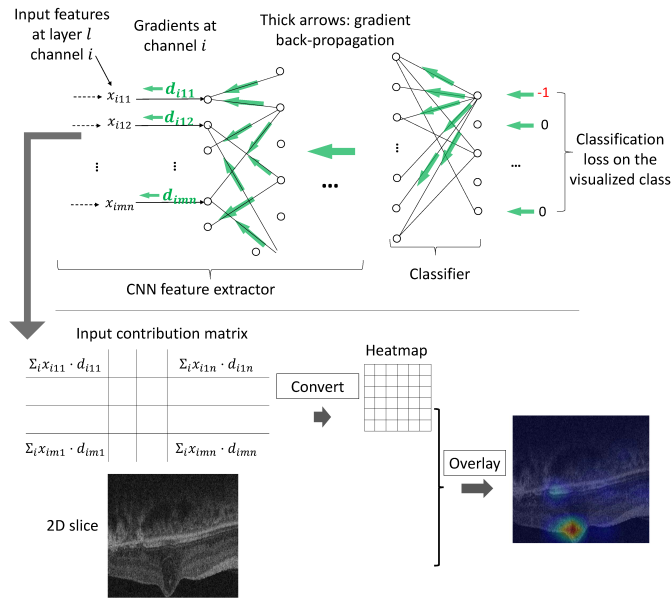}\caption{\label{fig:visflow}The flowchart of the gradient-based visualization
algorithm.}
\end{figure}

Denote the input slice as $R$. The heatmap algorithm proceeds
as follows:
\begin{enumerate}
\item Collect the gradient tensor at the $l$-th layer, denoted as ${d_{i11},\cdots,d_{imn}}$,
where $i$ indexes the feature channel, and $m,n$ indexes the height
and width, respectively. The gradient tensor are of the same size
of the input feature tensor to the $l$-th layer. The input feature
tensor to the $l$-th layer is denoted as $X={x_{i11},\cdots,x_{imn}}$.
Then perform an element-wise multiplication of the gradient tensor
and the input feature tensor, and sum out the feature channel dimension,
so as to get the input contribution matrix $T$, where the element
at the $i$-th row and the $j$-th column is:
\[
T_{hw}=\sum_{i}d_{ihw}\cdot x_{ihw};
\]
\item Set all the negative values in the input contribution matrix $T$
to 0, and scale to the values into the range $[0,255]$ and round
to integer values. Accordingly, $T$ is converted to a non-negative
contribution matrix, denoted as $P=\left(\begin{array}{ccc}
p_{11} & \cdots & p_{1n}\\
\vdots & \vdots & \vdots\\
p_{m1} & \cdots & p_{mn}
\end{array}\right)$. $P$ is interpolated to a heatmap $P^{*}$, which is of the same
size $W_{0}\cdot H_{0}$ as of the original image. $P^{*}$ highlights
some image areas, and the highlight color intensities are proportional
to the contributions of these areas to the final classification confidence
of class $c$.
\item A weighted sum of $P^{*}$ and $R$ is taken as the visualization
of the basis on which the classifier decides that the input image
is in class $c$: 
\[
H=0.6\cdot R+0.3\cdot P^{*}.
\]
\end{enumerate}
The data flow of the visualization algorithm is illustrated in Fig.\ref{fig:visflow}.
A heatmap of an OCT slice is presented in Fig.\ref{fig:vis}. We can
see that the Diabetic Macular Edema (DME) cyst is precisely localized
in the OCT slice.

\begin{center}
\includegraphics[scale=0.25]{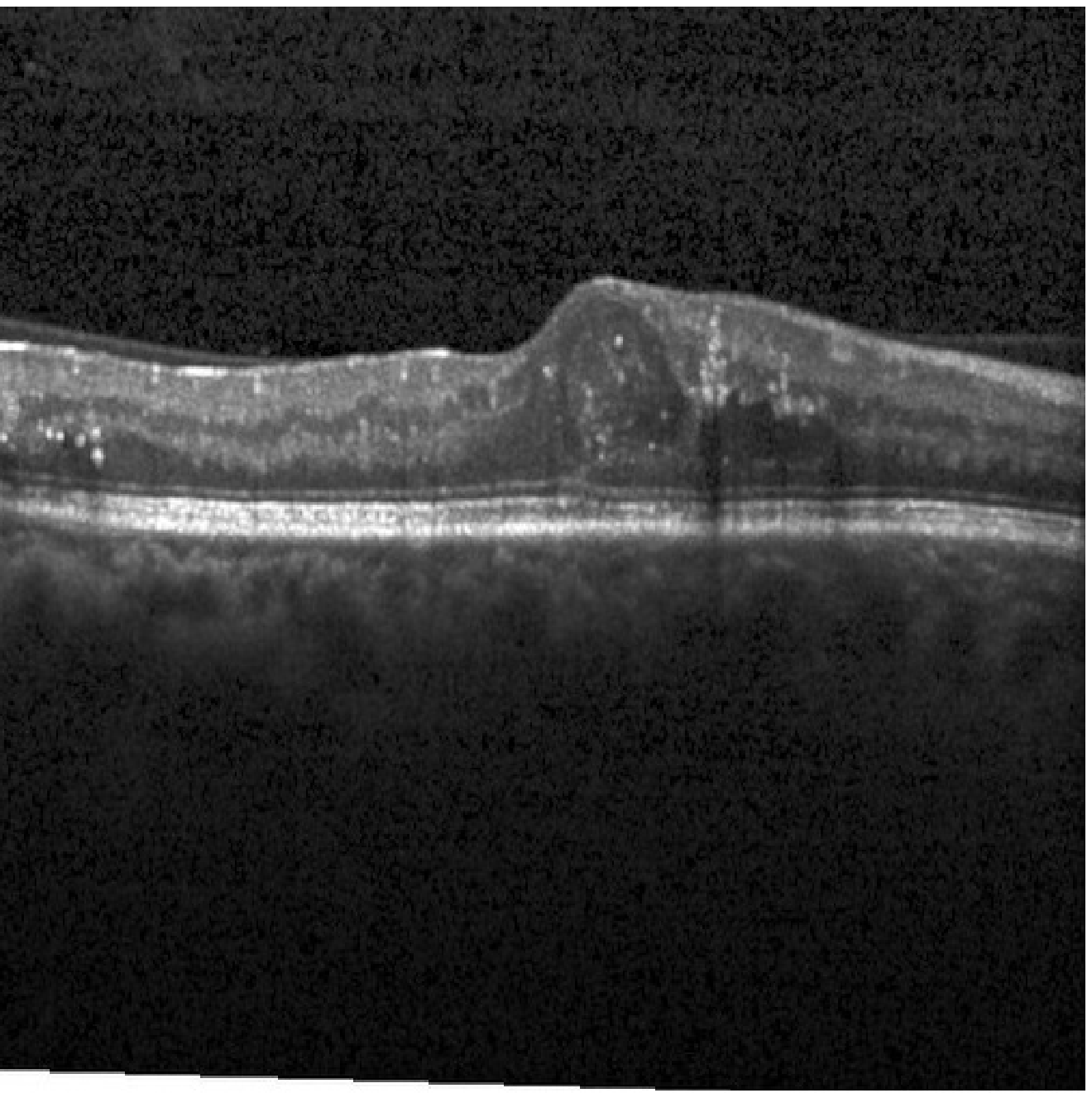}\includegraphics[scale=0.25]{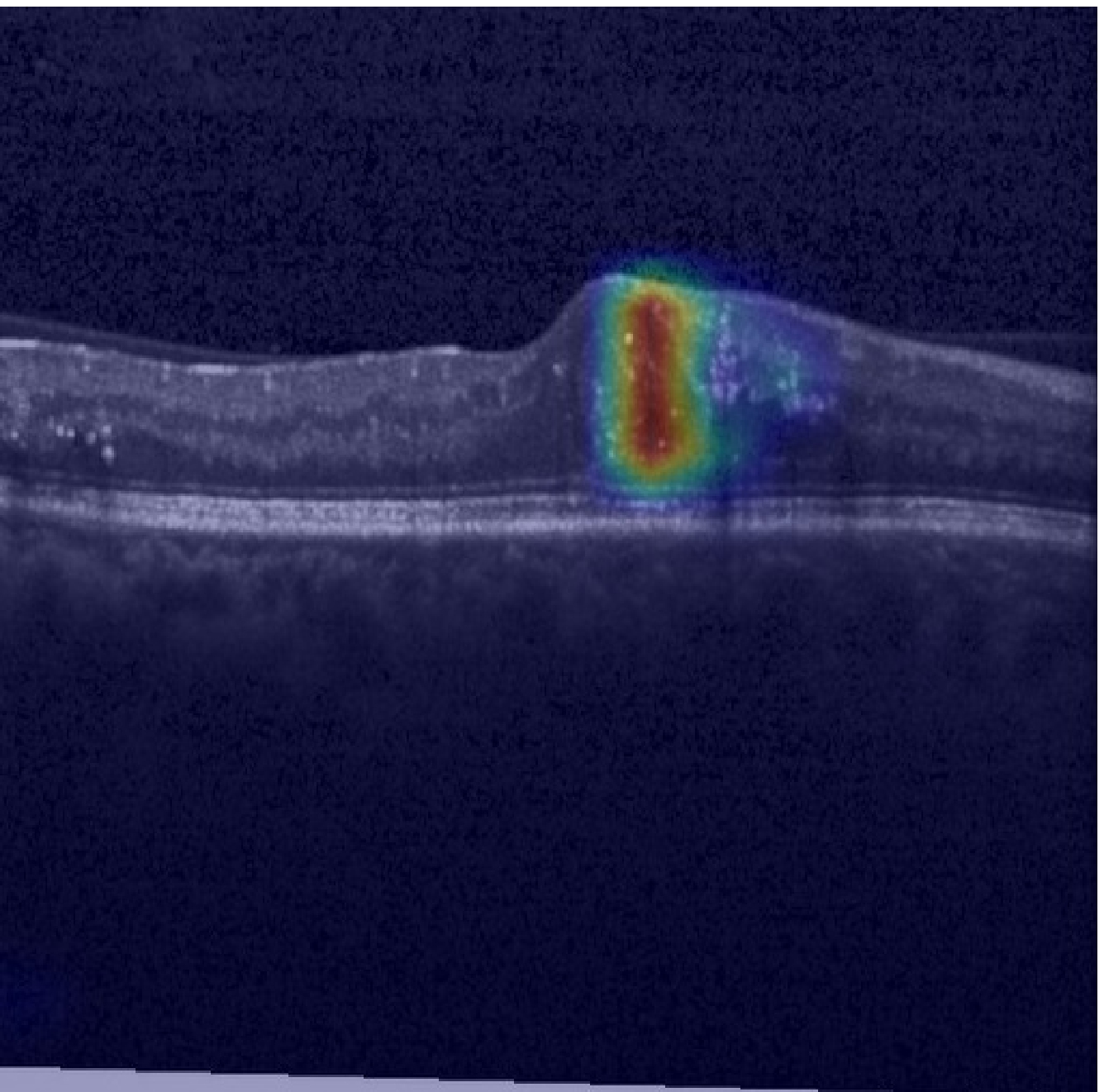}
\captionof{figure}{Visualization result of an OCT slice with DME.}\label{fig:vis}
\end{center}

\clearpage
\bibliographystyleappend{splncs04}
\bibliographyappend{references}

\end{document}